\documentclass{article}
\usepackage{spconf,amsmath,epsfig} 
\usepackage{amssymb} 
\usepackage{autobreak} 
\usepackage{booktabs}

\let\OLDthebibliography\thebibliography
\renewcommand\thebibliography[1]{
  \OLDthebibliography{#1}
  \setlength{\parskip}{0pt}
  \setlength{\itemsep}{0pt plus 0.3ex}
}

\begin{document}\sloppy

\def\x{{\mathbf x}}
\def\L{{\cal L}}

\title{Beyond MOS: Subjective Image Quality Score Preprocessing Method Based on Perceptual Similarity}
%
\name{Lei Wang, Desen Yuan}
\address{University of Electronic Science and Technology of China}

\maketitle

\begin{abstract}  

Image quality assessment often relies on raw opinion scores provided by subjects in subjective experiments, which can be noisy and unreliable. To address this issue, postprocessing procedures such as ITU-R BT.500, ITU-T P.910, and ITU-T P.913 have been standardized to clean up the original opinion scores. These methods use annotator-based statistical priors, but they do not take into account extensive information about the image itself, which limits their performance in less annotated scenarios. Generally speaking, image quality datasets usually contain similar scenes or distortions, and it is inevitable for subjects to compare images to score a reasonable score when scoring. Therefore, In this paper, we proposed Subjective Image Quality Score Preprocessing Method perceptual similarity Subjective Preprocessing (PSP), which exploit the perceptual similarity between images to alleviate subjective bias in less annotated scenarios. Specifically, we model subjective scoring as a conditional probability model based on perceptual similarity with previously scored images, called subconscious reference scoring. The reference images are stored by a neighbor dictionary, which is obtained by a normalized vector dot-product based nearest neighbor search of the images' perceptual depth features. Then the preprocessed score is updated by the exponential moving average (EMA) of the subconscious reference scoring, called similarity regularized EMA. Our experiments on multiple datasets (LIVE, TID2013, CID2013) show that this method can effectively remove the bias of the subjective scores. Additionally, Experiments prove that the Preprocesed dataset can improve the performance of downstream IQA tasks very well.
\end{abstract}
\section{Introduction}
\label{sec:intro}
\begin{figure}[!t]
    \centering
    \includegraphics[width=1\linewidth]{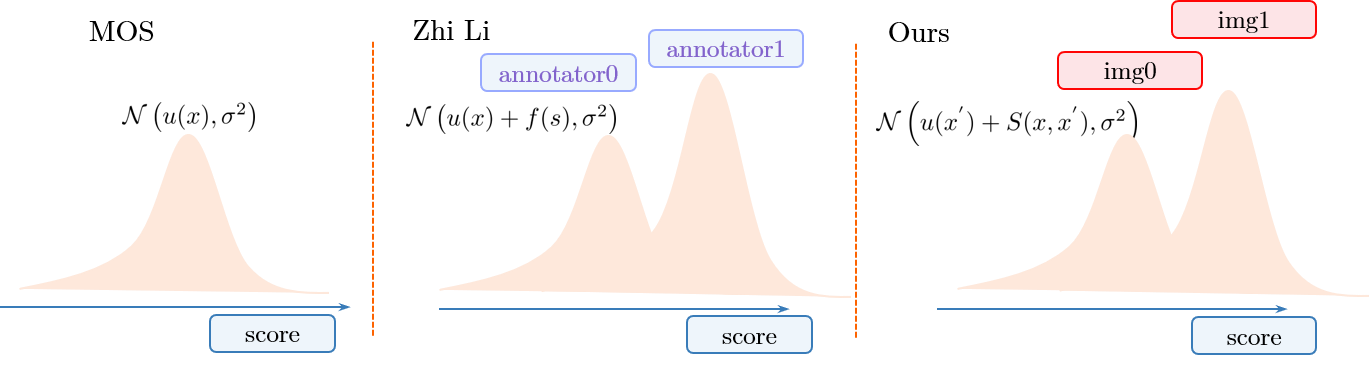}
    \caption{Left: traditional MOS model, Gaussian Distribution. Middle: model with annotator information added, Gaussian multimodal distribution. Right: model with image information added, Gaussian multimodal distribution. $u(x)$ is regarded as the true quality. $f(s)$ is a learnable bias of annotator $s$. $u(x^{'})$ is regarded as the true quality of the reference image. $S(x,x^{'})$ represents the score residual converted from the perceptual similarity between images.}
    \label{fig:moti}
    \vspace{-0.9em}
\end{figure}

Image quality assessment \cite{ma2017end,bosse2017deep,zhang2018blind,lin2019kadid,kundu2017large,virtanen2014cid2013,zaric2011vcl}, assessing the quality of an image is a crucial aspect of image processing research. Its goal is to enable computers to automatically analyze the features of an image and determine its strengths and weaknesses, such as the presence of distortion or other defects. The explosion of data has created a need for efficient processing of large amounts of image and video data. As a result, image quality assessment has become increasingly important for both practical applications and scientific research. It is typically divided into two categories: assessment with reference, and assessment without reference. Reference-based methods require a high-quality reference image to compare against, but this is not always feasible in practice. Therefore, no-reference assessment has become an active area of research in academia and industry.


To reduce the subjectivity of non-reference image quality assessment, this method requires manual labeling of image quality, collection of scores, and data processing to obtain the MOS value. However, there are various errors in human labeling. Currently, the processing methods only take into account the image and personnel numbers and only focus on the statistical results of the data. They ignore the feature information contained in the images themselves. Simply processing the labeling results statistically cannot eliminate the error of personnel labeling, but only provides a smooth process.


The core hypothesis of image quality assessment labeling is that two images with similar image quality have similar labeling scores. However, current MOS processing methods do not pay attention to the hidden information caused by image similarity, which is just the core of the image quality assessment task and does not have task specificity. For the first time, this article introduces image quality correlation to help solve Data postprocessing problems. For the task of image quality assessment without reference, the objective and unbiased tagging data is the most important for the training of the model.

Such a crowdsourcing way is costly and time-consuming. Some subjective score postprocessing methods have been proposed to replace MOS such as P910~\cite{installationsitu}, P913~\cite{union2016methods}, BT500~\cite{bt2020methodologies}. However, existing techniques for image quality assessment rely only on the collected scores and ignore the essential information contained in the original images. This can be problematic when the number of annotations is small or when only a single annotator is available. To address this issue, we propose a method named PSP-IQA for postprocessing IQA data based on subconscious reference scoring. It assumes that, When the human visual system makes a subjective quality assessment, it subconsciously gives a score based on a reference image. Specifically, we use the nearest neighbor search (NNS) to find a reference image based on perceptual similarity. Then similarity information in the dataset can help us to modify scores as a regularized item. It takes advantage of the hypothesis that similar image features should have similar scores, thus leveraging the implicit information in the dataset to correct the bias of subjective labeling. Although this regularization term of perceived similarity will limit the fraction away from his similar image. We still believe that the original marked score has a high degree of confidence, so we use EMA to fine-tune the original score by using the perceptual regularization term, which is called perceptual regularization fine-tuning. This simple and effective method is suitable for IQA data postprocessing and is particularly useful when only a limited number of annotations are available. The proposed perceptual similarity postprocessing method is proven to be effective in theory and experiment. We experimented with the IQA task (3 datasets). The results show that the proposed method reduces the adverse effect of subjective bias data on the model performance.


\begin{figure*}
	\centering
	\includegraphics[width=\linewidth]{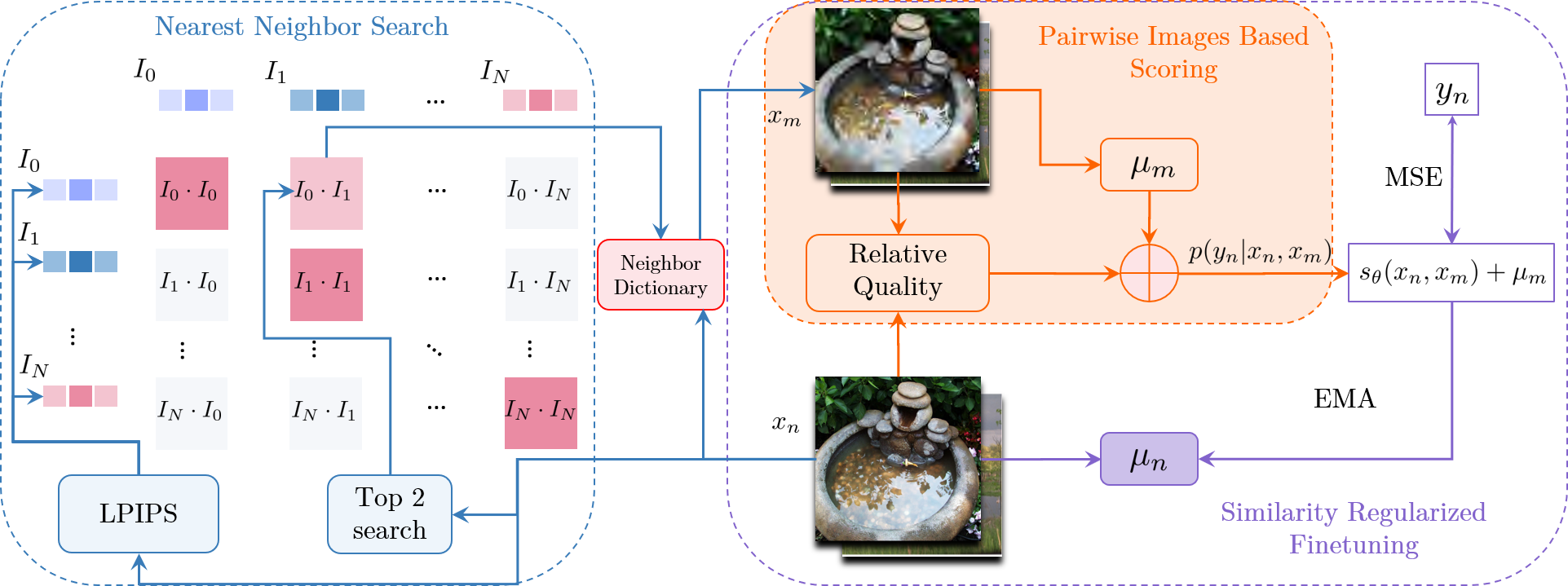} 
	\caption{The overall framework of the proposed method. When the marked score $y_{n}$ is biased large, the perceptual similarity score and the similar image estimation score can get the correct real score, thus correcting $y_{n}$.}
	\label{fig:arch}
\end{figure*}

\section{Related Work}



\noindent{\bf Image quality assessment.} 

For image quality assessment datasets \cite{lin2019kadid}, there are several acquisition and processing techniques, such as guaranteeing the consistency of the subjective assessment environment \cite{kundu2017large}, adding post-processing to the obtained data \cite{virtanen2014cid2013}, and eliminating outliers \cite{zaric2011vcl}. Without a doubt, however, several annotators are required for the large-scale subjective assessment datasets that are all amassed through crowdsourcing.

IQA tasks \cite{ma2017end,bosse2017deep,zhang2018blind}, which are frequently thought of as regression problems and have a regressive sigmoid head, have recently been the subject of DNN-based objective assessment algorithms that may perform better.
In addition, numerous alternative frameworks for subjective assessment tasks have also been developed using GAN \cite{ma2021blind}, VAE \cite{wang2020blind}, and transformers \cite{you2021transformer}.
For many novel visual tasks and scenarios, such as distorted images~\cite{lin2019kadid}, virtual reality~\cite{duan2018perceptual,sun2019mc360iqa}, light fields~\cite{Vamsi2017}, and dehazing images~\cite{min2018objective,min2019quality}, there is currently a high demand for fresh datasets. We try to reduce the number of annotations required by the data set by MOS post-processing method, while ensuring the reliability of the model.




\noindent{\bf Crowdsourced annotations and Noisy label.} 
Any complex function can be learned by an over-parameterized network from corrupted labels \cite{arpit2017closer,ghosh2017robust,tanaka2018joint,goldberger2016training,goldberger2016training,xia2021sample}. According to Zhang et al. \cite{zhang2021understanding}, DNNs can easily fit the entire training dataset with any corrupted label ratio, resulting in less generality on the test dataset. Robust loss functions~\cite{ghosh2017robust,tanaka2018joint,zhang2018generalized}, regularization~\cite{zhou2021learning}, robust network architecture~\cite{goldberger2016training}, sample selection~\cite{xia2021sample}, training strategy~\cite{hahn2019self,gotmare2018closer}, and other techniques are proposed to train deep networks in noisy environments.

Most methods, however, are designed around one-hot label properties such as classification task~\cite{zhou2021learning} sparsity and noise tolerance \cite{wang2019symmetric}. As a result, it cannot be directly applied to the subjective bias problem. Annotators for subjective assessment problems are frequently crowdsourced~\cite{zhuang2015leveraging,hube2019understanding,li2016crowdsourced}, and each person's score is biased against the ideal objective evaluator.


Furthermore, the International Telecommunication Union (ITU) and researchers have proposed a number of crowd-sourced data processing standards~\cite{bt2020methodologies,installationsitu,union2016methods} to eliminate subjective bias in MOS. However, these proposed methods do not take into account specific task information but instead rely on iterative fitting to remove subjective bias in the data, the applicability of different tasks is limited, and more data is required, making it difficult to apply under a single annotation.

\section{Methods}\label{sec:pl}

In previous papers, the main modeling method of subjective scoring can be expressed as the following formula:

\begin{equation}
p(y|x)=\mathrm{E}_{s|x} p(y|x,s)p(s|x)
\end{equation}

Rather than modeling the scoring $p(y|x)$ as a Gaussian distribution  $p(y|x) \sim  \mathcal{N}\left(u(x), {\sigma}^{2} \right)$, additional annotator information is introduced to model the conditional Gaussian distribution $p(y|x,s) \sim  \mathcal{N}\left(u(x)+f(s), {\sigma}^{2} \right)$. It is generally believed that annotators have the same bias for different images. $u(x)$ is regarded as the true quality.

Compared with introducing additional annotator information, we use the perceptual similarity of the image itself as a constraint. We consider the score of the mark to be related to a potential reference image when scoring.

\begin{equation}
p(y|x)= \mathrm{E}_{x^{'}|x} p(y|x,x^{'})p(x^{'}|x)
\end{equation}

where $p(y|x,x^{'}) \sim  \mathcal{N}\left(u(x^{'})+S(x,x^{'}), {\sigma}^{2} \right) $, $p(x^{'}|x)$ represents the probability distribution of the potential reference images when the annotator scores. $S(x,x^{'})$ represents the score residual converted from the perceptual similarity between images.

In fact, $p(x^{'}|x)$ is difficult to obtain because it is difficult to tell which image the annotator subconsciously compares with. 
It is natural to think that all images have an equal probability of being adopted by the annotator for reference. However, this would make us computationally unbearable.
For simplicity, we model it as a nearest-neighbor search model here. We think people always tend to compare with similar images. 
  
Given a noisy dataset $\{(x_n,y_n)\}^{N}_{i=1}$, maximizing the log-likelihood can be converted to minimizing the mean squared error (MSE).  
\begin{equation}
\begin{split}  \mathop{\arg\min}\limits_{\theta}\frac{1}{N} \sum_{n=1}^{N}(   y_{n}-(S( \boldsymbol{x}_{n},\text{NNS}(\boldsymbol{x}_{n}))+ u(\boldsymbol{x}_{n}^{'} ) )  )^{2},
\end{split}
\end{equation}  

In short, compared to traditional MOS $u(\boldsymbol{x}_{n})=y_{n}$, we added perceptual similarity as a regular term to fine-tune the final score.

Specifically, we use perceptual similarity LPIPS~\cite{zhang2018unreasonable} as a metric to search for the most similar images, resulting in NNS. The perceptual similarity score S is obtained by using the features of LPIPS through a scorer, which consists of a ResNet network.

When starting optimization, we initialize $u(\boldsymbol{x}_{n} )$ as $y_{n}$, by defining a score matrix $U \in \mathbb{R}^{1 \times N} $.
Iterate the network parameters via gradient descent and update the estimated true score by fine-tuning from $y$ using EMA:

\begin{equation}
\left\{
    \begin{array}{cc}
        u^{t+1}(\boldsymbol{x}_{n}) = \mathrm{EMA}(S^{t}( \boldsymbol{x}_{n},\boldsymbol{x}_{n}^{'})+ u^{t}(\boldsymbol{x}_{n}^{'} ) ,u^{t}(\boldsymbol{x}_{n})) & \\ 
        & \\
        \theta^{t+1}=\theta^{t}- \lambda  \Delta \theta
    \end{array}
\right.
\end{equation}
where $\Delta \theta$ obtained by backpropagation: 
\begin{equation}
\nabla_{\theta} \left\{\frac{1}{N} \sum_{n=1}^{N} (  y_{n}-(S^{t}( \boldsymbol{x}_{n},\text{NNS}(\boldsymbol{x}_{n}))+ u^{t}(\boldsymbol{x}_{n}^{'} ) ) )^2 \right\}.
\end{equation}

In the actual implementation, we set a warm-up time T for the perceptual similarity function S. Before the training epoch times T, we do not adjust the labels. After training the epoch times T times, we use EMA to fine-tune the score:

\begin{equation}
u^{t+1}(\boldsymbol{x}_{n}) =
\left\{
    \begin{array}{cc}
         \mathrm{EMA}(S^{t}( \boldsymbol{x}_{n},\boldsymbol{x}_{n}^{'})+ u^{t}(\boldsymbol{x}_{n}^{'} ) ,u^{t}(\boldsymbol{x}_{n})) & \\ 
        & \\
        y_{n} , \text{when}: t<T
    \end{array}
\right.
\end{equation}

\begin{table*} 
\centering
\caption{We compare the performance of our method with subjective preprocess methods and ablation study on bias label (bias rate 100\%) from image quality databases of LIVE, TID2013, and CID2013. We report SROCC, KROCC, and MSE results between the preprocessed quality scores and the true MOS provided by the database. We highlight the best results in bold font.}
\label{tab:compare-table}

\resizebox{\linewidth}{!}{
\begin{tabular}{l|cccc|cccc|cccc}
\toprule 
\toprule 
Datasets & \multicolumn{4}{c|}{LIVE} & \multicolumn{4}{c|}{TID} & \multicolumn{4}{c}{CID} \\ 
\midrule
Methods\textbackslash{}Metrics & SROCC & \multicolumn{1}{l}{PLCC} & \multicolumn{1}{l}{KROCC} & \multicolumn{1}{l|}{MSE} & SROCC & \multicolumn{1}{l}{PLCC} & \multicolumn{1}{l}{KROCC} & \multicolumn{1}{l|}{MSE} & SROCC & \multicolumn{1}{l}{PLCC} & \multicolumn{1}{l}{KROCC} & \multicolumn{1}{l}{MSE} \\ 
\midrule
MOS~\cite{streijl2016mean} & 0.7537 & 0.7510 & 0.5638 & 0.0314 & 0.7848 & 0.7956 & 0.5889 & 0.0160 & 0.6584 & 0.6753 & 0.4756 & 0.0588 \\
Zhi Li~\cite{li2020simple} & 0.7537 & 0.7510 & 0.5638 & 0.0314 & 0.7848 & 0.7956 & 0.5889 & 0.0160 & 0.6584 & 0.6753 & 0.4756 & 0.0588  \\
w/o score matrix & 0.7858 & 0.7598 & 0.5800 & 0.0281 & 0.8151 & 0.8274 & 0.6181 & 0.0248 & 0.6821 & 0.6904 & 0.4955 & 0.0434 \\
ours & \textbf{0.8501} & \textbf{0.8220} & \textbf{0.6593} & \textbf{0.0181} & \textbf{0.8329} & \textbf{0.8492} & \textbf{0.6387} & \textbf{0.0085} & \textbf{0.7374} & \textbf{0.7474} & \textbf{0.5436} & \textbf{0.0293} \\
\bottomrule
\bottomrule
\end{tabular}}
\end{table*}

\begin{table*} 
\centering
\caption{Performance comparison when the noise rate is set to different values. We conduct this test on the LIVE database and report the SROCC, PLCC, KROCC, and MSE results between the preprocessed quality scores and the true subjective scores of LIVE.}
\label{tab:rate-table}
\resizebox{\linewidth}{!}{
\begin{tabular}{l|cccc|cccc|cccc}
\toprule 
\toprule 
Datasets & \multicolumn{4}{c|}{LIVE} & \multicolumn{4}{c|}{TID} & \multicolumn{4}{c}{CID} \\ 
\midrule
Methods\textbackslash{}Metrics & SROCC & \multicolumn{1}{l}{PLCC} & \multicolumn{1}{l}{KROCC} & \multicolumn{1}{l|}{MSE} & SROCC & \multicolumn{1}{l}{PLCC} & \multicolumn{1}{l}{KROCC} & \multicolumn{1}{l|}{MSE} & SROCC & \multicolumn{1}{l}{PLCC} & \multicolumn{1}{l}{KROCC} & \multicolumn{1}{l}{MSE} \\ 
\midrule 
MOS rate=0.6 & 0.8437 & 0.8407 & 0.6886 & 0.0193 & 0.8565 & 0.8617 & 0.6991 & 0.0099 & 0.7675 & 0.7794 & 0.5988 & 0.0336 \\
Ours rate=0.6 & 0.8567 & 0.8339 & 0.6682 & 0.0190 & 0.8575 & 0.8713 & 0.6673 & 0.0083 & 0.8346 & 0.8350 & 0.6413 & 0.0201 \\
MOS rate=0.8 & 0.8231 & 0.8131 & 0.6451 & 0.0246 & 0.8133 & 0.8230 & 0.6308 & 0.0134 & 0.7004 & 0.7141 & 0.5149 & 0.0465 \\
Ours rate=0.8 & 0.8637 & 0.8429 & 0.6710 & 0.0169 & 0.8415 & 0.8550 & 0.6475 & 0.0082 & 0.7578 & 0.7595 & 0.5599 & 0.0277 \\
\bottomrule
\bottomrule
\end{tabular}}
\end{table*}

\section{Experiments}\label{sec:exp}

\label{sec:experiments}
In this section, we first describe the experimental setups, including datasets, assessment criteria, and network architecture details. Then we compare the performance of PSP with other preprocessing methods. We next conduct a series of ablation studies to identify the contribution of the key components of PSP. Finally, we also present some visualization cases.

\subsection{Datasets and Settings}
\label{sec:nmt}
\noindent{\bf Datasets.} 
Since there are no existing image quality assessment datasets to measure preprocessing performance with specific annotation information. We selected existing popular quality assessment datasets to generate data with subjective bias, including LIVE \cite{sheikh2006statistical} and TID2013 \cite{ponomarenko2015image} and popular quality assessment datasets CID2013 \cite{virtanen2014cid2013} with specific annotation information.

If the dataset contains annotation information, we randomly sample the labeled scores of each image and average them to obtain a MOS with subjective bias as the training set. The original MOS is used as the test set. If the data set does not contain annotation information, we use the original MOS of each image as the mean, and the variance takes the given variance of the data set (or sets it to 0.2) as a Gaussian distribution to obtain a biased MOS. 

\noindent{\bf Evaluation Criteria.} 
The Pearson Linear Correlation Coefficient (PLCC), Spearman Rank Order Correlation Coefficient (SROCC), Kendall rank-order correlation coefficient (KROCC), and MSE are used to measure performance, as in previous work. 



\subsection{Detailed Implementation}
Since perceptual similarity models have recently demonstrated great capability. We employ LPIPS and ResNet-50 as the backbone. The images in the training set are resized to 320, and randomly cropped to 320. The images in the test set are fed directly into our model with no data augmentation. All of this is done with the assumption that preprocessed images have the same score as the original. We use ResNet50 as the proposed method's backbone network based on the general configuration of network structure in IQA fields. The model's hyperparameter settings include learning rate = 0.01, SGD as the optimizer, epoch = 10, and training batch size = 16. The model was trained using a single GeForce RTX 3090 GPU.

\subsection{Performance assessment and Comparison} 
In this section, experiments within individual standard IQA databases are conducted to evaluate the effectiveness of PSP. We discuss how to use single subjective labels to achieve performance under the labels obtained from many annotators. As the question has never been explored, we select the typical subjective assessment models, namely ResNet-50~\cite{he2016deep} and LPIPS~\cite{zhang2018unreasonable}, and then test the accuracy of the models when using single subjective labels.  To simulate the scenario of single labels, we replace all the labels in the datasets (except for CID2013). In other words, the bias label rate is 100 percent. Single subjective labels are generated via Gaussian Sampling from the labels processed by several annotators. Real single labels are obtained on CID2013 datasets, and the labels processed by several annotators are replaced randomly with the labels processed by a single annotator.

The experimental results in table \ref{tab:compare-table} demonstrate the correlation between the score obtained by a small amount of score (just need one score in \ref{tab:compare-table}) and the real Ground Truth. The closer the first three indexes (SROCC, PLCC, and KROCC) are to one, the closer they are to Ground Truth, and the smaller the last is (MSE), the smaller the margin of error. Note that the comparison here is a comparison of the accuracy of the pre-processing method, and the quality of the labeling directly affects the performance of the model training. Compared with other mainstream crowdsourced subjective scores processing methods, such as MOS and Zhi Li's, the proposed method is superior to others in SROCC, PLCC, KROCC, and MSE. The results are shown in Table \ref{tab:compare-table}. The accuracy of the models is significantly lower when trained with single subjective labels than when trained with the labels processed in the original dataset (standard MOS). It indicates that the bias from the single subjective labels exerts a great negative impact on the accuracy of the models. However, the accuracy remains at similar levels when trained by the subjective labels processed by several annotators and when trained by the single subjective labels. When trained by single subjective labels, the proposed model is more accurate than other subjective biased models. we have the following observations. 

\subsection{Ablation studies}
\subsubsection{Different Noise Rates}
Because there may be different annotations of labels for a single image in the case of true labeling, this paper calls it the noise rate. In order to simulate the scene and verify the effectiveness of the proposed PSP-IQA method, experiments are carried out under the noise rate of 0.6 and 0.8 respectively. The results are shown in Table \ref{tab:rate-table}, the proposed method outperforms the MOS method at different noise rates (e. g. 0.6,0.8,1), which verifies the universality of PSP-IQA.

\subsubsection{Different EMA Weights}
To evaluate the performance of the proposed model, we performed ablation experiments with respect to the EMA parameters of the model. In order to test the performance changes under different EMA weights, interval ablation experiments ranging from 0.2 to 1.0 were set up. The results are shown in Table \ref{tab:weight-table}.

\begin{table} 
\centering
\caption{The performance impact of the dataset after the PSP subjective quality score preprocessing method on the quality assessment task. We conduct this test on the LIVE database and report SROCC, PLCC, KROCC, and MSE results between quality rating performance and LIVE true subjective scores. The quality evaluation model uses NIMA. The subjective bias scale was set to 1.0 and the bias variance to 0.2. We found that the scoring after PSP subjective quality score preprocessing can endow the quality evaluation model with better performance.}
 
\begin{tabular}{l|c|c|c|c}
\toprule 
\toprule 
 & SROCC & \multicolumn{1}{l|}{PLCC} & \multicolumn{1}{l|}{KROCC} & \multicolumn{1}{l}{MSE} \\ \midrule
MOS-GT & 0.9317 & 0.8988 & 0.7790 & 0.0112 \\ \midrule
MOS    & 0.8704 & 0.8653 & 0.6877 & 0.0238 \\ \midrule
PSP    & 0.9202 & 0.8966 & 0.7582 & 0.0178 \\
\bottomrule
\bottomrule
\end{tabular} 

\label{tab:iqa-table}
\end{table}

\begin{table} 
\centering
\caption{Performance comparison when the EMA Weights is set to different values. We conduct this test on the LIVE database and report the SROCC, PLCC, KROCC, and MSE results between the preprocessed quality scores and the true subjective scores of LIVE. We highlight the best results in bold}
\label{tab:weight-table} 
\begin{tabular}{l|c|c|c|c}
\toprule 
\toprule 
 & SROCC & \multicolumn{1}{l|}{PLCC} & \multicolumn{1}{l|}{KROCC} & \multicolumn{1}{l}{MSE} \\ \midrule
EMA=0.2 & 0.4752 & 0.4324 & 0.3205 & 0.1247 \\ \midrule
EMA=0.4 & 0.5802 & 0.5492 & 0.4005 & 0.0758 \\ \midrule
EMA=0.8 & 0.7800 & 0.7484 & 0.5808 & 0.0271 \\ \midrule
EMA=0.9 & \textbf{0.8501} & \textbf{0.8220} & \textbf{0.6593} & \textbf{0.0181} \\ \midrule
EMA=1.0 & 0.7537 & 0.7510 & 0.5638 & 0.0314 \\
\bottomrule
\bottomrule
\end{tabular}
\end{table}

 


\subsubsection{Different variance and $T$} 
\begin{figure}
    \centering
    \includegraphics[width=1\linewidth]{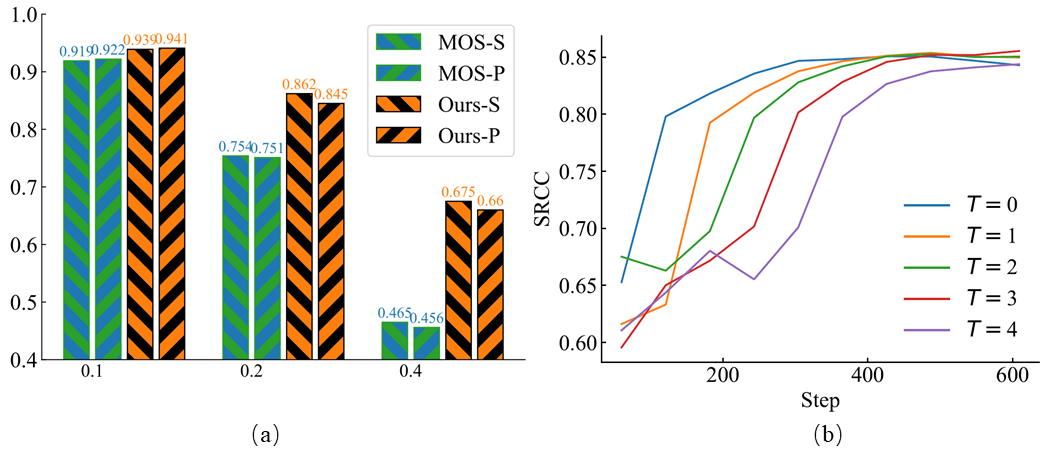}
    \caption{(a) ResNet-50: Accuracies on LIVE with different variance $\sigma$.(b) ResNet-50: Accuracies on LIVE with different $T$.}
    \label{fig:sth}
    \vspace{-0.9em}
\end{figure}

We performed ablation experiments to $\sigma$ of the Gaussian distribution $p(y|x) \sim  \mathcal{N}\left(u(x)+f(s), {\sigma}^{2} \right)$ and to the $T$ of the EMA. The results are shown in Figure \ref{fig:sth}. As shown in figure \ref{fig:sth} (a), the proposed PSP-IQA method is superior to the MOS method in different variance $\sigma$, and the larger the variance $\sigma$ is, the more the performance gap between PSP-IQA and MOS method can be reflected, and the validity of the proposed method is verified. As shown in figure \ref{fig:sth} (b), the convergence rate of PSP-IQA is different under different T, but the final convergence precision is close, which shows that the proposed method is insensitive to iterative parameters and robust.

\begin{figure}
    \centering
    \includegraphics[width=1\linewidth]{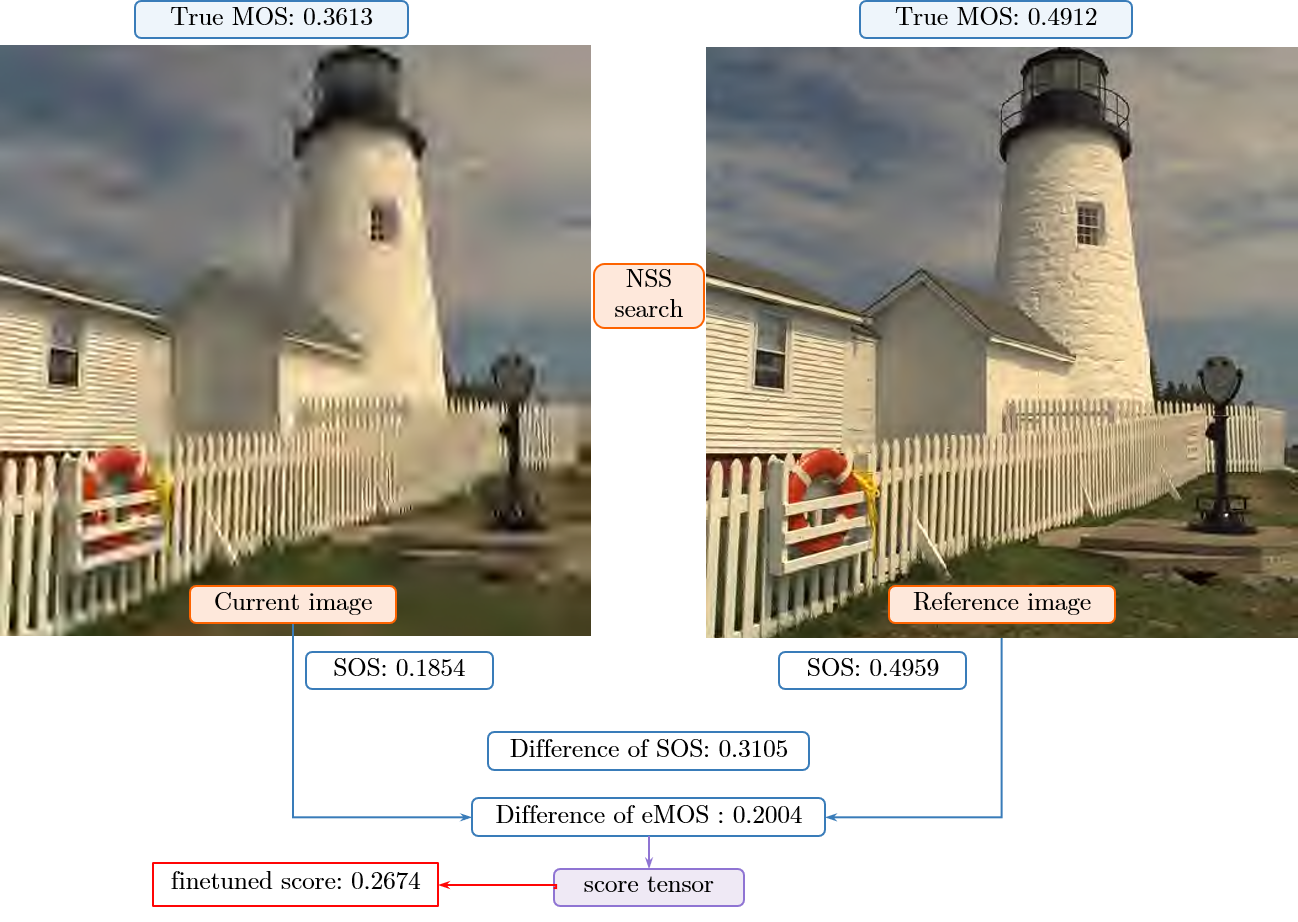}
    \caption{Given a biased MOS (below the image), our method searches for images with similar perceptual similarities via NSS and gets the finetune score.}
    \label{fig:case}
    \vspace{-0.9em}
\end{figure}

\subsection{Case Study}
We conduct a case study to show the effectiveness of LPR-IQA. As show in Fig. \ref{fig:case}, we found that calculating the perceptual similarity score between the two found a large gap with the actual annotated similarity score. In this way, we estimate a score matrix to fine-tune the score so that the score is more consistent with the actual distortion of the image. It is worth noting that our method fine-tunes the score by weighing the content of all images in the dataset and the distribution of scores, rather than the two images shown. The marked score of the image we selected is very different from the real score, and the corresponding reference image searched by NSS is also the same. There is a gap between our perceptual similarity score ($0.2905$) and the annotated similarity score ($0.5441$). However, the perceptual similarity score is correlated with the whole dataset with higher confidence. Based on the perceptual similarity score, we were able to fine-tune the image score.

\section{Conclusions and Future Work}

This paper proposes a new IQA-perceptual similarity processing method, which is a simple and effective image-perceptual similarity-based IQA data preprocessing method PSP-IQA, suitable for IQA tasks. PSP-IQA alleviates the subjective bias problem when there are few annotators. This method can consider not only the existing annotation labels, but also the perceptual similarity relationship of images in the data set. In our approach, we assume that the subjective annotation process is a scoring process based on latent reference images, relative to the perceived similarity of another image. We propose a conditional probability model based on LPISP and nearest neighbor search (NSS) to model the subjective scoring process. And fine-tune the existing labels based on EMA. On the real dataset CID2013 and the popular IQA datasets LIVE and TID, it is verified that our method can effectively alleviate the bias of subjective scoring.

\bibliographystyle{IEEEbib}
\bibliography{icme2023template}

\end{document}